\documentclass[11pt,a4paper]{article}
\usepackage[hyperref]{naaclhlt2019}
\pdfoutput=1

\aclfinalcopy 

\usepackage{url}
\usepackage{setspace}
\usepackage[]{algorithm2e}
\usepackage{times}
\usepackage{latexsym}
\usepackage{amsmath}
\usepackage{multirow}
\usepackage{graphicx}
\usepackage{amssymb}
\usepackage{amsfonts}
\usepackage{todonotes}
\usepackage{fancyvrb}
\usepackage{tikz}
\usepackage{xcolor}
\usepackage{color}
\usepackage{float}
\usepackage{dcolumn}
\usepackage{graphicx}
\usepackage[english]{babel}
\usepackage[bottom]{footmisc}
\usepackage{capt-of}
\usepackage{fixltx2e}

\usepackage{pgfplots}
\usepackage{pgfplotstable} 
\pgfplotsset{compat=newest}

\title{A Capsule Network-based Embedding Model for Knowledge Graph Completion and Search Personalization}

\author{Dai Quoc Nguyen${}^{1}$, Thanh Vu${}^{2}$, Tu Dinh Nguyen${}^{1}$, Dat Quoc Nguyen${}^{3}$, Dinh Phung${}^{1}$ \\
${}^{1}$Monash University, Australia; ${}^{3}$The University of Melbourne, Australia\\ ${}^{2}$The Australian e-Health Research Centre, CSIRO, Australia\\
${}^{1}${\tt{{\{dai.nguyen,tu.dinh.nguyen,dinh.phung\}@monash.edu}}} \\
${}^{2}${\tt{{thanh.vu@csiro.au}}}; ${}^{3}${\tt{{dqnguyen@unimelb.edu.au}}}
}

\begin{document}
\maketitle

\begin{abstract}

In this paper, we introduce an embedding model, named CapsE, exploring a capsule network to model relationship triples \textit{(subject, relation, object)}. Our CapsE represents each triple as a 3-column matrix where each column vector represents the embedding of an element in the triple. This 3-column matrix is then fed to a convolution layer where multiple filters are operated to generate different feature maps. These feature maps are reconstructed into corresponding capsules which are then routed to another capsule to produce a continuous vector. The length of this vector is used to measure the plausibility score of the triple. Our proposed CapsE obtains better performance than previous state-of-the-art embedding models for knowledge graph completion on two benchmark datasets WN18RR and FB15k-237, and outperforms strong search personalization baselines on SEARCH17.


\end{abstract}

\section{Introduction}
Knowledge graphs (KGs) containing relationship triples \textit{(subject, relation, object)}, denoted as \textit{(s, r, o)}, are the useful resources for many NLP and especially information retrieval applications such as semantic  search and question answering \citep{8047276}. 
However, large knowledge graphs, even containing billions of triples, are still incomplete, i.e., missing a lot of valid triples \citep{West:2014}. 
Therefore, much research efforts have focused on the knowledge graph completion task which aims to predict missing triples in KGs, i.e., predicting whether a triple not in KGs is likely to be valid or not \citep{bordes2011learning,NIPS2013_5071,NIPS2013_5028}. 
To this end, many embedding models have been proposed to learn vector representations for entities (i.e., \textit{subject}/head entity and \textit{object}/tail entity) and relations in KGs, and obtained state-of-the-art results as summarized by \citet{NickelMTG15} and \citet{Nguyen2017}.
These embedding models score triples \textit{(s, r, o)}, such that valid triples have higher plausibility scores than invalid ones \citep{bordes2011learning,NIPS2013_5071,NIPS2013_5028}. 
For example, in the context of KGs, the score for \textit{(Melbourne, cityOf, Australia)} is higher than the score for \textit{(Melbourne, cityOf, United Kingdom)}.
 
Triple modeling is  applied not only to the KG completion, but also for other tasks which can be formulated as a triple-based prediction problem. An example is in search personalization, one would aim to tailor search results to each specific user based on the user's personal interests and preferences \citep{Teevan2005,Teevan2009,Bennett2012,HarveyB2013,Vu2015,vu2017search}.
Here the triples can be formulated as \textit{(submitted query, user profile, returned document)} and used to re-rank documents returned to a user given an input query, by employing an existing KG embedding method  such as TransE \citep{NIPS2013_5071}, as proposed by \citet{vu2017search}.
Previous studies have shown the effectiveness of modeling triple for either KG completion or search personalization.
However, there has been no single study investigating the performance on both  tasks.

Conventional embedding models, such as TransE \citep{NIPS2013_5071}, DISTMULT \citep{Yang2015} and ComplEx \citep{Trouillon2016}, use addition, subtraction or simple multiplication operators, thus only capture the linear relationships between entities.
Recent research has raised interest in applying deep neural networks to triple-based prediction problems.
For example, \citet{Nguyen2018} proposed ConvKB---a convolutional neural network (CNN)-based model for KG completion and achieved state-of-the-art results. Most of KG embedding models are constructed to modeling entries at the same dimension of the given triple, where presumably each dimension captures some relation-specific attribute of entities.
To the best of our knowledge, however, none of the existing models has a ``deep'' architecture for modeling the entries in a triple at the same dimension.

\citet{sabour2017dynamic} introduced capsule networks (CapsNet) that employ capsules (i.e., \textit{each capsule is a group of neurons}) to capture entities in images and then uses a routing process to specify connections from capsules in a layer to those in the next layer.
Hence CapsNet could encode the intrinsic spatial relationship between a part and a whole constituting viewpoint invariant knowledge that automatically generalizes to novel viewpoints.
Each capsule accounts for capturing variations of an object or object part in the image, which can be efficiently visualized. Our high-level hypothesis is that embedding entries at the same dimension of the triple also have these variations, although it is not straightforward to be visually  examined. 

To that end, we introduce CapsE to explore a novel application of CapsNet on triple-based data for two problems: KG completion and search personalization. Different from the traditional modeling design of CapsNet where capsules are constructed by splitting feature maps, we use capsules to model the entries at the same dimension in the entity and relation embeddings.
In our CapsE, $\boldsymbol{v}_s$, $\boldsymbol{v}_r$ and $\boldsymbol{v}_o$ are unique $k$-dimensional embeddings of $s$, $r$ and $o$, respectively. 
The embedding triple [$\boldsymbol{v}_s$, $\boldsymbol{v}_r$, $\boldsymbol{v}_o$] of \textit{(s, r, o)} is fed to the convolution layer where multiple filters of the same $1\times3$ shape are repeatedly operated over every row of the matrix to produce  $k$-dimensional feature maps.
Entries at the same dimension from all feature maps are then encapsulated into a capsule.
Thus, \textit{each capsule can encode many characteristics in the embedding triple to represent the entries at the corresponding dimension}.
These capsules are then routed to another capsule which outputs a continuous vector whose length is used as a score for the triple.
Finally, this score is used to predict whether the triple \textit{(s, r, o)} is valid or not.

In summary, our main contributions from this paper are as follows:

$\bullet$ We propose an embedding model CapsE using the capsule network \citep{sabour2017dynamic} for modeling relationship triples.
To our best of knowledge, our work is the first consideration of exploring the capsule network to knowledge graph completion and search personalization.

$\bullet$ We evaluate our  CapsE for knowledge graph completion on two benchmark datasets WN18RR \citep{Dettmers2017} and FB15k-237 \citep{toutanova-chen:2015:CVSC}. CapsE obtains the best mean rank on WN18RR and the highest mean reciprocal rank and highest Hits@10 on FB15k-237.

$\bullet$ We restate the prospective strategy of expanding the triple embedding models to improve the ranking quality of the search personalization systems.
We adapt our model to search personalization and evaluate on SEARCH17 \citep{vu2017search} -- a dataset of the web search query logs.  Experimental  results show  that our CapsE achieves the new state-of-the-art results with significant improvements over strong baselines.


\begin{figure*}[ht]
\centering
\includegraphics[width=0.95\textwidth]{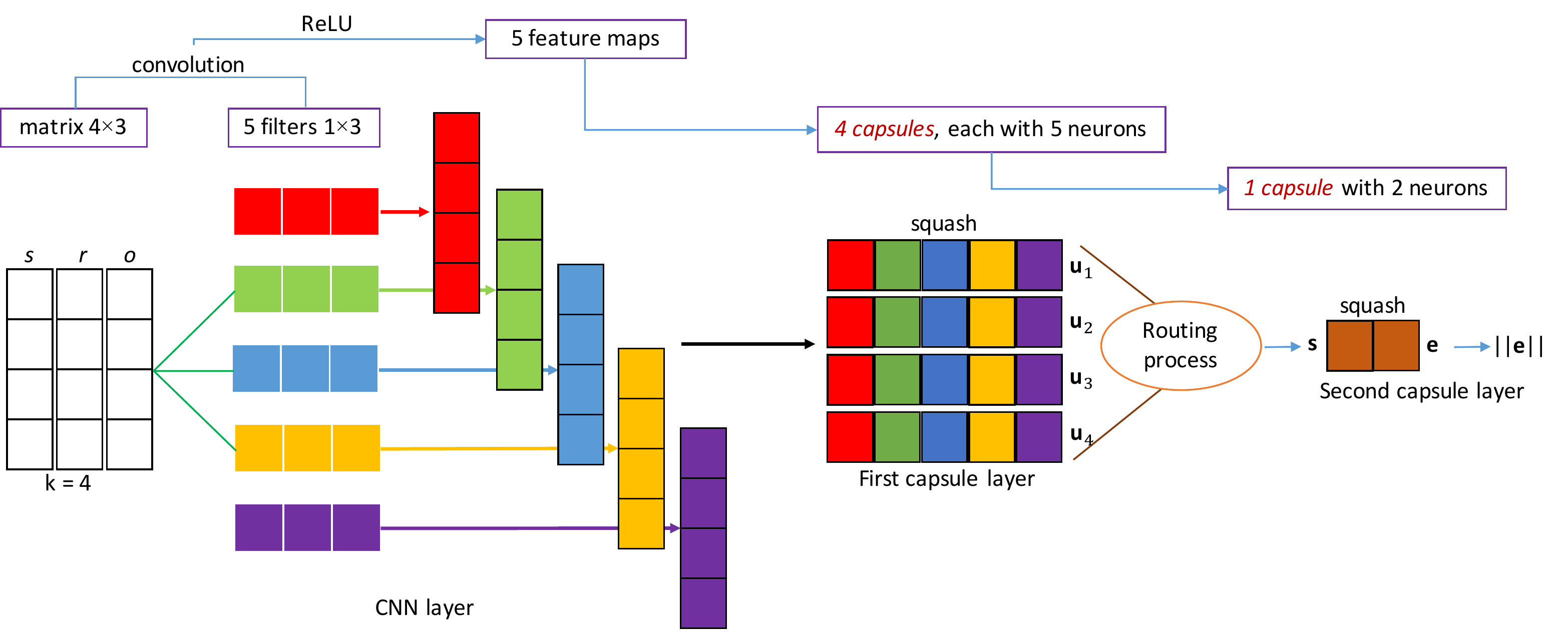}
\caption{An example illustration of our CapsE with $k=4$, $\mathsf{N}=5$, and $d = 2$.}
\label{fig:model}
\end{figure*}

\section{The proposed CapsE}
\label{sec:ourmodel}

Let $\mathcal{G}$ be a collection of valid factual triples in the form of \textit{(subject, relation, object)} denoted as \textit{(s, r, o)}. 
Embedding models aim to define a \textit{score function} giving a score for each triple, such that valid triples receive higher scores than invalid triples.

We denote $\boldsymbol{v}_s$, $\boldsymbol{v}_r$ and $\boldsymbol{v}_o$ as the $k$-dimensional embeddings of $s$, $r$ and $o$, respectively.
In our proposed CapsE, we follow \citet{Nguyen2018} to view each embedding triple [$\boldsymbol{v}_s$, $\boldsymbol{v}_r$, $\boldsymbol{v}_o$] as a matrix $\boldsymbol{A} = [\boldsymbol{v}_s,\boldsymbol{v}_r,\boldsymbol{v}_o] \in \mathbb{R}^{k\times3}$, and denote $\boldsymbol{A}_{i,:} \in \mathbb{R}^{1\times3}$ as the $i$-th row of $\boldsymbol{A}$. 
We use a filter $\boldsymbol{\omega} \in \mathbb{R}^{1\times3}$ operated on the convolution layer.
This filter $\boldsymbol{\omega}$ is repeatedly operated over every row of $\boldsymbol{A}$ to generate a feature map $\boldsymbol{\mathsf{q}} = [\mathsf{q}_1, \mathsf{q}_2, ..., \mathsf{q}_k] \in \mathbb{R}^{k}$, in which
$\mathsf{q}_i = g\left(\boldsymbol{\omega} \cdot{\boldsymbol{A}_{i,:}} + b\right)$ where $\cdot$ denotes a dot product, 
$b \in \mathbb{R}$ is a bias term and $g$ is a non-linear activation function such as ReLU. 
Our model uses multiple filters $\in \mathbb{R}^{1\times3}$ to generate feature maps. 
We denote $\bold{\Omega}$ as the set of filters and $\mathsf{N} = \mid\bold{\Omega}\mid$ as the number of filters, thus we have $\mathsf{N}$ $k$-dimensional feature maps, for which each feature map can capture one single characteristic among entries at the same dimension.

We build our CapsE with two single capsule layers for a simplified architecture.
In the first layer, we construct $k$ capsules, wherein entries at the same dimension from all feature maps are encapsulated into a corresponding capsule. 
Therefore, each capsule can capture many characteristics among the entries at the corresponding dimension in the embedding triple. 
These characteristics are generalized into one capsule in the second layer which produces a vector output whose length is used as the score for the triple.

The first capsule layer consists of $k$ capsules, for which each capsule $i \in \{1, 2, ..., k\}$ has a vector output $\boldsymbol{\mathsf{u}}_{i} \in \mathbb{R}^{\mathsf{N}\times1}$.
Vector outputs $\boldsymbol{\mathsf{u}}_{i}$ are multiplied by weight matrices $\textbf{W}_{i} \in \mathbb{R}^{d\times\mathsf{N}}$ to produce vectors $\hat{\boldsymbol{\mathsf{u}}}_{i} \in \mathbb{R}^{d\times1}$ which are summed to produce a vector input $\boldsymbol{\mathsf{s}} \in \mathbb{R}^{d\times1}$ to the capsule in the second layer.
The capsule then performs the non-linear squashing function to produce a vector output $\boldsymbol{\mathsf{e}} \in \mathbb{R}^{d\times1}$:
\begin{eqnarray}
\boldsymbol{\mathsf{e}} = \mathsf{squash}\left(\boldsymbol{\mathsf{s}}\right) \ \ \ ;  \ \ \ \boldsymbol{\mathsf{s}} = \sum_{i} c_{i}\hat{\boldsymbol{\mathsf{u}}}_{i} \ \ \ ;  \ \ \ \hat{\boldsymbol{\mathsf{u}}}_{i} = \textbf{W}_{i} \boldsymbol{\mathsf{u}}_{i} \nonumber
\end{eqnarray}
where $\mathsf{squash}\left(\boldsymbol{\mathsf{s}}\right) = \frac{\|\boldsymbol{\mathsf{s}}\|^2}{1 + \|\boldsymbol{\mathsf{s}}\|^2}\frac{\boldsymbol{\mathsf{s}}}{\|\boldsymbol{\mathsf{s}}\|}$, and $c_{i}$ are coupling coefficients determined by the routing process as presented in Algorithm \ref{alg:routing}. 
Because there is one capsule in the second layer, we make only one difference in the routing process proposed by \citet{sabour2017dynamic}, for which we apply the $\mathsf{softmax}$ in a direction from all capsules in the previous layer to each of capsules in the next layer.\footnote{The $\mathsf{softmax}$ in the original routing process proposed by \citet{sabour2017dynamic} is applied in another direction from each of capsules in the previous layer to all capsules in the next layer.}

\begin{algorithm}[ht]
\DontPrintSemicolon
\SetAlgoVlined
\setstretch{1.25}

\For{all capsule i $\in$ the first layer}{
    $b_{i} \leftarrow$ 0 
}
\For{$\mathsf{iteration}$ = 1, 2, ..., m}{
    $\boldsymbol{\mathsf{c}} \leftarrow \mathsf{softmax}\left(\boldsymbol{\mathsf{b}}\right)$ 
    
    $\boldsymbol{\mathsf{s}} \leftarrow \sum_{i} c_{i}\hat{\boldsymbol{\mathsf{u}}}_{i}$
    
    $\boldsymbol{\mathsf{e}} = \mathsf{squash}\left(\boldsymbol{\mathsf{s}}\right)$
    
    \For{all capsule i $\in$ the first layer}{
        $b_{i} \leftarrow b_{i} + \hat{\boldsymbol{\mathsf{u}}}_{i}\cdot\boldsymbol{\mathsf{e}}$ 
}
}
\caption{The routing process is extended from \citet{sabour2017dynamic}.}
\label{alg:routing}
\end{algorithm}

We illustrate our proposed model in Figure \ref{fig:model} where embedding size: $k=4$, the number of filters: $\mathsf{N}=5$, the number of neurons within the capsules in the first layer is equal to $\mathsf{N}$, and the number of neurons within the capsule in the second layer: $d = 2$. The length of the vector output $\boldsymbol{\mathsf{e}}$ is used as the score for the input triple. 

Formally, we define the score function $f$ for the triple $(s, r, o)$ as follows:
\begin{equation}
f\left(s,r,o\right) = \|\mathsf{capsnet}\left(g\left([\boldsymbol{v}_s,\boldsymbol{v}_r,\boldsymbol{v}_o]\ast\bold{\Omega}\right)\right)\| \nonumber
\label{equa:model}
\end{equation}
where the set of filters $\bold{\Omega}$ is shared parameters in the convolution layer; $\ast$ denotes a convolution operator; and $\mathsf{capsnet}$ denotes a capsule network operator.
We use the Adam optimizer \citep{kingma2014adam} to train CapsE by minimizing the loss function \citep{Trouillon2016,Nguyen2018} as follows:
\begin{align}
\mathcal{L} & =  \sum_{\substack{(s, r, o) \in \{\mathcal{G} \cup \mathcal{G}'\}}} \log\left(1 + \exp\left(- t_{(s, r, o)} \cdot f\left(s,r,o\right)\right)\right) \nonumber
 \label{equal:objfunc}
  \end{align}
  \vspace{-10pt}
\begin{equation*}
\text{in which, } t_{(s, r, o)} = \left\{ 
  \begin{array}{l}
  1\;\text{for } (s, r, o)\in\mathcal{G}\\
 -1\;\text{for } (s, r, o)\in\mathcal{G}'
  \end{array} \right.
\end{equation*}
here $\mathcal{G}$  and $\mathcal{G}'$  are collections of valid and invalid triples, respectively.  $\mathcal{G}'$ is generated by corrupting valid triples in $\mathcal{G}$.


\section{Knowledge graph completion evaluation }\label{ssec:kbc}
In the knowledge graph completion task \citep{NIPS2013_5071}, the goal is to predict a missing entity given a relation and another entity, i.e, inferring a head entity $s$ given $(r, o)$ or inferring a tail entity $o$ given $(s, r)$. 
The results are calculated based on ranking the scores produced by the score function $f$ on test triples.

\subsection{Experimental setup} 
\noindent\textbf{Datasets:} 
We use two recent benchmark datasets WN18RR \citep{Dettmers2017} and FB15k-237 \citep{toutanova-chen:2015:CVSC}. These two datasets are created to avoid  reversible relation problems, thus the prediction task becomes more realistic and hence more challenging \citep{toutanova-chen:2015:CVSC}.
Table \ref{tab:datasets} presents the statistics of WN18RR and FB15k-237.

\begin{table}[!htb]
\centering
\resizebox{7.75cm}{!}{
\setlength{\tabcolsep}{0.4em}
\begin{tabular}{l|lllll}
\hline
\bf Dataset &  \#E & \#R  & \multicolumn{3}{l}{\#Triples in train/valid/test}\\
\hline
WN18RR & 40,943 & 11 & 86,835 & 3,034 & 3,134\\
FB15k-237 & 14,541 & 237 & 272,115 & 17,535 & 20,466\\
\hline
\end{tabular}
}
\caption{Statistics of the experimental  datasets. \#E is the number of entities. \#R is the number of relations.}
\label{tab:datasets}
\end{table}

\noindent\textbf{Evaluation protocol:}
Following \citet{NIPS2013_5071}, for each valid test triple $(s, r, o)$, we replace either $s$ or $o$ by each of all other entities to create a set of corrupted triples.
We use the ``{Filtered}'' setting protocol \citep{NIPS2013_5071}, i.e., not taking  any corrupted triples that appear in the KG into accounts.
We rank the valid test triple and corrupted triples in descending order of their scores.
We employ evaluation metrics: mean rank (MR), mean reciprocal rank (MRR) and Hits@10 (i.e., the proportion of the valid test triples ranking in top {10} predictions).
Lower MR, higher MRR or higher Hits@{10} indicate better performance. Final scores on the test set are reported for the model obtaining the
highest Hits@10 on the validation set.

\noindent\textbf{Training protocol:}
We use the common Bernoulli strategy \citep{AAAI148531,AAAI159571} when sampling invalid triples. 
For WN18RR, \citet{yuvalpinterm3gm}\footnote{
\citet{yuvalpinterm3gm} considered WN18RR and evaluated their M3GM model only for 7 relations as they employed the inverse rule model \citep{Dettmers2017} for 4 remaining symmetric relations. 
Regarding a fair comparison to other models, we use the M3GM implementation released by \citet{yuvalpinterm3gm} to re-train and re-evaluate the M3GM model for all 11 relations. 
We thank \citet{yuvalpinterm3gm} for their assistance running their code.
} 
found a strong evidence to support the necessity of a WordNet-related semantic setup, in which they averaged pre-trained word embeddings for word surface forms within the WordNet to create synset embeddings, and then used these synset embeddings to initialize entity embeddings for training their TransE association model.
We follow this evidence in using the pre-trained 100-dimensional Glove word embeddings \citep{pennington2014glove} to train a TransE model on WN18RR.

We employ the TransE and ConvKB implementations provided by \citet{NguyenNAACL2016} and \citet{Nguyen2018}. 
For ConvKB, we use a new process of training up to 100 epochs and monitor the Hits@10 score after every 10 training epochs to choose optimal hyper-parameters with the Adam initial learning rate in $ \{1e^{-5}, 5e^{-5}, 1e^{-4}\}$ and the number of filters $\mathsf{N}$ in $\{50, 100, 200, 400\}$.
We obtain the highest Hits@10 scores on the validation set when using N= 400 and the initial learning rate $5e^{-5}$ on WN18RR; and N= 100 and the initial learning rate $1e^{-5}$ on FB15k-237.

Like in ConvKB, we use the same pre-trained entity and relation embeddings produced by TransE to initialize entity and relation embeddings in our CapsE for both WN18RR and FB15k-237 ($k = 100$).
We set the batch size to 128, the number of neurons within the capsule in the second capsule layer to 10 ($d = 10$), and the number of iterations in the routing algorithm $m$ in $\{1, 3, 5, 7\}$.
We run CapsE up to 50 epochs and monitor the Hits@10 score after each 10 training epochs to choose optimal hyper-parameters.
The highest Hits@10 scores for our CapsE on the validation set are obtained when using $m = 1$, $\mathsf{N} = 400$
and the initial learning rate at $1e^{-5}$ on WN18RR; 
and $m = 1$, $\mathsf{N} = 50$
and the initial learning rate at $1e^{-4}$ on FB15k-237.

\subsection{Main experimental results}

\begin{table*}[!htb]
\centering
\begin{tabular}{l|lll|lll}
\hline
\multirow{2}{*}{\bf Method}& \multicolumn{3}{c|}{\bf WN18RR} & \multicolumn{3}{c}{\bf FB15k-237}\\
\cline{2-7}
 & MR & MRR & H@10 & MR & MRR & H@10 \\
\hline
DISTMULT \citep{Yang2015} & 5110 & {0.425} & 49.1 & \underline{254} & 0.241 & 41.9 \\
ComplEx \citep{Trouillon2016} & 5261 & \textbf{0.444} & 50.7 & 339 & 0.247 & 42.8\\
ConvE \citep{Dettmers2017} & 4187 & \underline{0.433} & 51.5 & \textbf{244} & 0.325 & 50.1\\
KBGAN \citep{Cai2017} & -- & 0.213 & 48.1 & -- & 0.278 & 45.8 \\
M3GM \citep{yuvalpinterm3gm} & 1864 & 0.311 & 53.3 & -- & -- & --\\
\hline
TransE \citep{NIPS2013_5071} & \underline{743}$^{\star}$ & 0.245$^{\star}$ & \underline{56.0}$^{\star}$ & 347 & 0.294 & 46.5 \\
ConvKB \citep{Nguyen2018} & 763$^{\star}$ & 0.253$^{\star}$ & \textbf{56.7}$^{\star}$ & \underline{254}$^{\star}$ & \underline{0.418}$^{\star}$ & \underline{53.2}$^{\star}$ \\
\hline
Our \textbf{CapsE} & \textbf{719} & {0.415} & \underline{56.0} & 303 & \textbf{0.523} & \textbf{59.3} \\
\hline
\end{tabular}
\caption{Experimental results on the WN18RR and FB15k-237 test sets. Hits@10 (H@10) is reported in \%. Results of DISTMULT, ComplEx and ConvE are taken from \citet{Dettmers2017}.
Results of TransE on FB15k-237 are taken from \citet{Nguyen2018}.
Our CapsE \textbf{Hits@1} scores are 33.7\% on WN18RR and \textbf{48.9}\% on FB15k-237.
Formulas of MRR and Hits@1 show a strong correlation, so using Hits@1 does not really reveal any additional information for this task.
The best score is in \textbf{bold}, while the second best score is in \underline{underline}. 
$\star$ denotes {\textit{our new results}} for TransE and ConvKB, which are better than those published by \citet{Nguyen2018}.
}
\label{tab:results}
\end{table*}

\begin{figure*}[!htb]
\centering
\centering
\resizebox{15.5cm}{!}{
\begin{tikzpicture}
\begin{axis}[
	title = Predicting $head$, 
    ybar,
    enlarge x limits=0.25,
    legend style={at={(0.275,1)},
                anchor=north,legend columns=2},
    ylabel={Hits@10},
    symbolic x coords={1-1, 1-M, M-1, M-M},
    xtick=data,
    ymin=0,ymax=100,
    nodes near coords=\rotatebox{90}{\scriptsize\pgfmathprintnumber\pgfplotspointmeta},
    ]
\addplot coordinates {(1-1,47.40) (1-M,47.72) (M-1,59.78) (M-M,57.28)};
\addplot coordinates {(1-1,51.56) (1-M,57.15) (M-1,49.51) (M-M,48.85)};
\legend{CapsE,ConvKB}
\end{axis}
\end{tikzpicture}
\hspace{0.35cm}
\begin{tikzpicture}
\begin{axis}[
	title = Predicting $tail$, 
    ybar,
    enlarge x limits=0.25,
    legend style={at={(0.275,1)},
                anchor=north,legend columns=2},
    ylabel={Hits@10},
    symbolic x coords={1-1, 1-M, M-1, M-M},
    xtick=data,
    ymin=0,ymax=100,
    nodes near coords=\rotatebox{90}{\scriptsize\pgfmathprintnumber\pgfplotspointmeta},
    ]
\addplot coordinates {(1-1,47.92) (1-M,17.48) (M-1,80.26) (M-M,58.59)};
\addplot coordinates {(1-1,50.00) (1-M,11.37) (M-1,83.58) (M-M,53.46)};
\legend{CapsE,ConvKB}
\end{axis}
\end{tikzpicture}

\begin{tikzpicture}
\begin{axis}[
	title = Predicting $head$, 
    ybar,
    enlarge x limits=0.25,
    legend style={at={(0.275,1)},
                anchor=north,legend columns=2},
    ylabel={MRR},
    symbolic x coords={1-1, 1-M, M-1, M-M},
    xtick=data,
    ymin=0,ymax=0.85,
    nodes near coords=\rotatebox{90}{\scriptsize\pgfmathprintnumber\pgfplotspointmeta},
    ]
\addplot coordinates {(1-1,0.447) (1-M,0.281) (M-1,0.594) (M-M,0.512)};
\addplot coordinates {(1-1,0.465) (1-M,0.368) (M-1,0.476) (M-M,0.373)};
\legend{CapsE,ConvKB}
\end{axis}
\end{tikzpicture}
\hspace{0.35cm}
\begin{tikzpicture}
\begin{axis}[
	title = Predicting $tail$, 
    ybar,
    enlarge x limits=0.25,
    legend style={at={(0.275,1)},
                anchor=north,legend columns=2},
    ylabel={MRR},
    symbolic x coords={1-1, 1-M, M-1, M-M},
    xtick=data,
    ymin=0,ymax=0.85,
    nodes near coords=\rotatebox{90}{\scriptsize\pgfmathprintnumber\pgfplotspointmeta},
    ]
\addplot coordinates {(1-1,0.446) (1-M,0.168) (M-1,0.693) (M-M,0.516)};
\addplot coordinates {(1-1,0.455) (1-M,0.096) (M-1,0.719) (M-M,0.394)};
\legend{CapsE,ConvKB}
\end{axis}
\end{tikzpicture}

}
\caption{Hits@10 (in \%) and MRR on the FB15k-237 test set w.r.t each relation category.}
\label{fig:hits10relationtype}
\end{figure*}
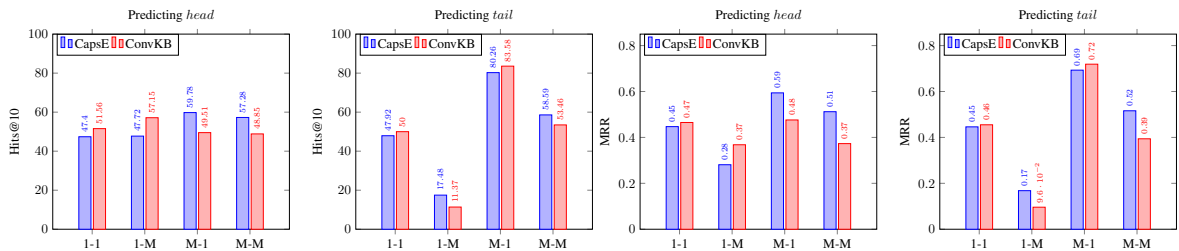

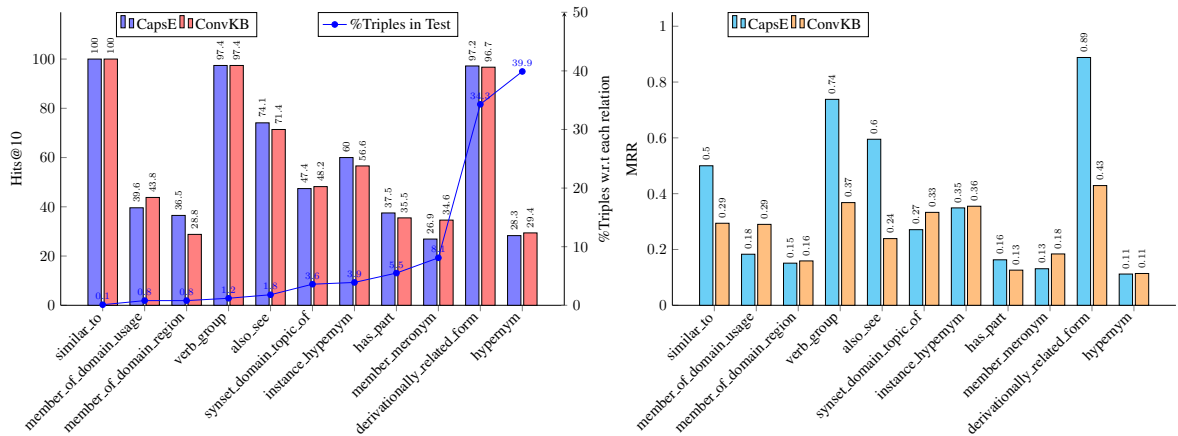
\begin{figure*}[!htb]
\centering
\resizebox{15.5cm}{!}{
\begin{tikzpicture}
\pgfplotsset{lineplot/.style={blue,mark=*,sharp plot,line legend}}
  \begin{axis}[
	width=0.65\textwidth,
	x=1.05cm,
	enlarge x limits={abs=0.5},
	bar width=9.5pt,
  	ybar,
  	axis x line* = bottom,
    axis y line* = left,
    enlarge x limits=0.1,
    legend style={at={(0.25,1)},
                anchor=north,legend columns=2},
    ylabel={Hits@10},
    xtick=data,
    ymin=0,ymax=119,
    nodes near coords=\rotatebox{90}{\scriptsize\pgfmathprintnumber\pgfplotspointmeta},
    x tick label style={rotate=45,anchor=east},
    symbolic x coords = {
    	similar\_to,
		member\_of\_domain\_usage,
		member\_of\_domain\_region,
		verb\_group,
		also\_see,
		synset\_domain\_topic\_of,
		instance\_hypernym,
		has\_part,
		member\_meronym,
		derivationally\_related\_form,
		hypernym, 
    },
  ]
  
  \addplot [fill=blue!50] coordinates {
  		(hypernym,28.3)
		(derivationally\_related\_form,97.2)
		(member\_meronym,26.9)
		(has\_part,37.5)
		(instance\_hypernym,60.0)
		(synset\_domain\_topic\_of,47.4)
		(also\_see,74.1)
		(verb\_group,97.4)
		(member\_of\_domain\_region,36.5)
		(member\_of\_domain\_usage,39.6)
		(similar\_to,100)
  
  };
  \addplot [fill=red!50] coordinates {
  		(hypernym,29.4)
		(derivationally\_related\_form,96.7)
		(member\_meronym,34.6)
		(has\_part,35.5)
		(instance\_hypernym,56.6)
		(synset\_domain\_topic\_of,48.2)
		(also\_see,71.4)
		(verb\_group,97.4)
		(member\_of\_domain\_region,28.8)
		(member\_of\_domain\_usage,43.8)
		(similar\_to,100)  
  };
  \legend{CapsE,ConvKB}
  \end{axis}
  
      \begin{axis}[
  	width=0.65\textwidth,
	x=1.05cm,
	enlarge x limits={abs=0.5},
	bar width=9.5pt,
  	ybar,
  	hide x axis,
  	axis y line=right,
    enlarge x limits=0.1,
    legend style={at={(0.65,1)},
                anchor=north,legend columns=2},
    ylabel={\%Triples w.r.t each relation},
    xtick=data,
    ymin=0,ymax=50,
    nodes near coords={\scriptsize\pgfmathprintnumber\pgfplotspointmeta},
    x tick label style={rotate=45,anchor=east},
    symbolic x coords = {		
		similar\_to,
		member\_of\_domain\_usage,
		member\_of\_domain\_region,
		verb\_group,
		also\_see,
		synset\_domain\_topic\_of,
		instance\_hypernym,
		has\_part,
		member\_meronym,
		derivationally\_related\_form,
		hypernym, 
    },
  ]
  \addplot [lineplot,color=blue] coordinates {
  		(hypernym,39.9)
		(derivationally\_related\_form,34.3)
		(member\_meronym,8.1)
		(has\_part,5.5)
		(instance\_hypernym,3.9)
		(synset\_domain\_topic\_of,3.6)
		(also\_see,1.8)
		(verb\_group,1.2)
		(member\_of\_domain\_region,0.8)
		(member\_of\_domain\_usage,0.8)
		(similar\_to,0.1)  
  };
  \legend{\%Triples in Test}
  \end{axis}
  
\end{tikzpicture}

\begin{tikzpicture}
\pgfplotsset{lineplot/.style={blue,mark=*,sharp plot,line legend}}
  \begin{axis}[
	width=0.65\textwidth,
	x=1.05cm,
	enlarge x limits={abs=0.5},
	bar width=9.5pt,
  	ybar,
  	axis x line* = bottom,
    axis y line* = left,
    enlarge x limits=0.1,
    legend style={at={(0.25,1)},
                anchor=north,legend columns=2},
    ylabel={MRR},
    xtick=data,
    ymin=0.000,ymax=1.050,
    nodes near coords=\rotatebox{90}{\scriptsize\pgfmathprintnumber\pgfplotspointmeta},
    x tick label style={rotate=45,anchor=east},
    symbolic x coords = {
    	similar\_to,
		member\_of\_domain\_usage,
		member\_of\_domain\_region,
		verb\_group,
		also\_see,
		synset\_domain\_topic\_of,
		instance\_hypernym,
		has\_part,
		member\_meronym,
		derivationally\_related\_form,
		hypernym, 
    },
  ]
  
  \addplot [fill=cyan!50] coordinates {
  		(hypernym,0.112)
		(derivationally\_related\_form,0.888)
		(member\_meronym,0.131)
		(has\_part,0.163)
		(instance\_hypernym,0.349)
		(synset\_domain\_topic\_of,0.271)
		(also\_see,0.595)
		(verb\_group,0.738)
		(member\_of\_domain\_region,0.151)
		(member\_of\_domain\_usage,0.183)
		(similar\_to,0.500)
  
  };
  \addplot [fill=orange!50] coordinates {
  		(hypernym,0.114)
		(derivationally\_related\_form,0.429)
		(member\_meronym,0.184)
		(has\_part,0.126)
		(instance\_hypernym,0.355)
		(synset\_domain\_topic\_of,0.333)
		(also\_see,0.239)
		(verb\_group,0.368)
		(member\_of\_domain\_region,0.159)
		(member\_of\_domain\_usage,0.290)
		(similar\_to,0.294)  
  };
  \legend{CapsE,ConvKB}
  \end{axis}
  
\end{tikzpicture}
}
\caption{Hits@10 and MRR on the WN18RR test set w.r.t each relation. The right y-axis is the percentage of triples corresponding to relations.}
\label{fig:eachrelationWN18RRresults}
\end{figure*}

Table \ref{tab:results} compares the experimental results of our CapsE with previous state-of-the-art published results, using the same evaluation protocol.
Our CapsE performs better than its closely related CNN-based model ConvKB on both experimental datasets (except Hits@10 on WN18RR and MR on FB15k-237), especially on FB15k-237 where our CapsE gains significant improvements of $0.523 - 0.418 = 0.105$ in MRR (which is about 25.1\% relative improvement), and $59.3\% - 53.2\% = 6.1$\% absolute improvement in Hits@10. 
Table \ref{tab:results} also shows that our CapsE obtains the best MR score on WN18RR and the highest MRR and Hits@10 scores on FB15k-237.

Following \citet{NIPS2013_5071}, for each relation $r$ in FB15k-237, we calculate the averaged number $\eta_s$ of head entities per tail entity and the averaged number $\eta_o$ of tail entities per head entity. 
If $\eta_s < $1.5 and $\eta_o < $1.5, $r$ is categorized  one-to-one  (1-1).
If $\eta_s < $1.5 and $\eta_o \geq $1.5, $r$ is categorized  one-to-many (1-M).
If $\eta_s \geq $1.5 and $\eta_o < $1.5, $r$ is categorized  many-to-one (M-1).
If $\eta_s \geq $1.5 and $\eta_o \geq $1.5, $r$ is categorized  many-to-many (M-M). 
As a result, 17, 26, 81 and 113 relations are labelled  1-1, 1-M, M-1 and M-M, respectively. 
And 0.9\%, 6.3\%, 20.5\% and 72.3\% of the test triples in FB15k-237 contain 1-1, 1-M, M-1 and M-M relations, respectively.

Figure \ref{fig:hits10relationtype} shows the Hits@10 and MRR results for predicting head and tail entities w.r.t each relation category on FB15k-237.
CapsE works better than ConvKB in predicting entities on the ``side M'' of triples (e.g., predicting \textit{head} entities in M-1 and M-M; and predicting \textit{tail} entities in 1-M and M-M), while ConvKB performs better than CapsE in predicting entities on the ``side 1'' of triples (i.e., predicting \textit{head} entities in 1-1 and 1-M; and predicting \textit{tail} entities in 1-1 and M-1).

Figure \ref{fig:eachrelationWN18RRresults} shows the Hits@10 and MRR scores w.r.t each relation on WN18RR.
$also\_see$, $similar\_to$, $verb\_group$ and $derivationally\_related\_form$ are symmetric relations which can be considered as M-M relations.
Our CapsE also performs better than ConvKB on these 4 M-M relations.
Thus, results shown in Figures \ref{fig:hits10relationtype} and  \ref{fig:eachrelationWN18RRresults} are consistent.
These also imply that our CapsE would be a potential candidate for applications which contain many M-M relations such as search personalization. 

We see that the length and orientation of each capsule in the first layer can also help to model the important entries in the corresponding dimension, thus CapsE can work well on the ``side M'' of triples where entities often appear less frequently than others appearing in the ``side 1'' of triples.
Additionally, existing models such as DISTMULT, ComplEx and ConvE can perform well for entities with high frequency, but may not for rare entities with low frequency.
These are reasons why our CapsE can be considered as the best one on FB15k-237 and it outperforms most existing models on WN18RR.

\begin{table}[!t]
\centering
\begin{tabular}{c|ccccc}
\hline
$m$ & 10 & 20 & 30 & 40 & 50\\
\hline
1 & \textbf{48.37} & \textbf{52.60} & \textbf{53.14} & \textbf{53.33} & \textbf{53.21}\\
3 & 47.78 & 52.34 & 52.93 & 52.99 & 52.86\\
5 & 47.03 & 52.25 & 45.80 & 45.99 & 45.76\\
7 & 40.46 & 45.36 & 45.79 & 45.85 & 45.93\\
\hline
\end{tabular}
\caption{Hits@10 on the WN18RR validation set with $\mathsf{N} = 50$ and the initial learning rate at $1e^{-5}$ w.r.t each number of iterations in the routing algorithm $m$ and each 10 training epochs.} 
\label{tab:effectofrouting}
\end{table}

\textbf{Effects of routing iterations:} We study how the number of routing iterations affect the performance.
Table \ref{tab:effectofrouting} shows the Hits@10 scores on the WN18RR validation set for a comparison w.r.t each number value of the routing iterations and epochs with the number of filters $\mathsf{N} = 50$ and the Adam initial learning rate at $1e^{-5}$.
We see that the best performance for each setup over each 10 epochs is obtained by setting the number $m$ of routing iterations to 1. 
This indicates the opposite side for knowledge graphs compared to images. 
In the image classification task, setting the number $m$ of iterations in the routing process higher than 1 helps to capture the relative positions of entities in an image (e.g., eyes, nose and mouth) properly.
In contrast, this property from images may be only right for the 1-1 relations, but not for the 1-M, M-1 and M-M relations in the KGs because of the high variant of each relation type (e.g., symmetric relations) among different entities.

\section{Search personalization application}
\label{subsec:sp}

Given a \textit{user}, a submitted \textit{query} and the \textit{documents} returned by a search system for that query, our approach is to re-rank the returned documents so that the more relevant documents should be ranked higher.  
Following \citet{vu2017search}, we represent the relationship between the submitted query, the user and the returned document as a \textit{(s, r, o)}-like triple \textit{(query, user, document)}. The triple captures how much interest a user puts on a document given a query. 
Thus, we can evaluate the effectiveness of our CapsE for the search personalization task.

\subsection{Experimental setup}

\noindent\textbf{Dataset:} 
We use the SEARCH17 dataset \citep{vu2017search} of query logs of 106 users collected by a large-scale web search engine.
A log entity consists of a user identifier, a query, top-10 ranked documents returned by the search engine and clicked documents along with the user's dwell time. \citet{vu2017search} constructed short-term (session-based) user profiles and used the profiles to personalize the returned results.
They then employed the SAT criteria \citep{FoxE2005} to identify whether a returned document is relevant from the query logs as either a clicked document with a dwell time of at least 30 seconds or the last clicked document in a search session (i.e., a SAT click). 
After that, they assigned a $relevant$ label to a returned document if it is a SAT click and also assigned $irrelevant$ labels to the remaining top-10 documents.
The rank position of the $relevant$ labeled documents is used as the ground truth to evaluate the search performance before and after re-ranking.


The dataset was uniformly split into the training, validation and test sets. This split is for the purpose of using historical data in the training set to predict new data in the test set \citep{vu2017search}. The training, validation and test sets consist of 5,658, 1,184 and 1,210 relevant (i.e., valid) triples; and 40,239, 7,882 and 8,540 irrelevant (i.e., invalid) triples, respectively.

\noindent\textbf{Evaluation protocol:} 
Our CapsE is used to re-rank the original list of documents returned by a search engine as follows: (i) We train our model and employ the trained model to calculate the score for each  $(s, r, o)$ triple. (ii) We then sort the scores in the descending order to obtain a new ranked list.
To evaluate the performance of our proposed model, we use two standard evaluation metrics: mean reciprocal rank (MRR) and Hits@1.\footnote{We re-rank the list of top-10 documents returned by the search engine, so Hits@10 scores are same for all models.}
For each metric, the higher value indicates better ranking performance.

We compare CapsE with the following baselines using the same experimental setup: (\textbf{1}) SE: The original rank is returned by the search engine. (\textbf{2}) CI \citep{Teevan2011}: This baseline uses a personalized navigation method based on previously clicking returned documents. (\textbf{3}) SP \citep{Bennett2012,Vu2015}: A search personalization method makes use of the session-based user profiles. (\textbf{4}) 
Following \citet{vu2017search}, we use TransE as a strong baseline model for the search personalization task.  Previous work shows that the well-known embedding model TransE, despite its simplicity, obtains very competitive results for the knowledge graph completion \citep{lin-EtAl:2015:EMNLP1,nickel2016holographic,Trouillon2016,Nguyen2016,Nguyen2018}. (\textbf{5}) The CNN-based model ConvKB is the most closely related model to our CapsE.

\noindent\textbf{Embedding initialization:} We follow \citet{vu2017search}  to initialize user profile, query and document embeddings for the baselines TransE and ConvKB, and our CapsE.

We train a LDA topic model \citep{Blei2003} with 200 topics only on the \textit{relevant} documents (i.e., SAT clicks) extracted from the query logs. We then use the trained LDA model to infer the probability distribution over topics for every returned document. We use the topic proportion vector of each document as its document embedding (i.e. $k=200$). In particular, the $z^{th}$ element ($z = 1,2,...,k$) of the vector embedding for document $d$ is: $\boldsymbol{v}_{d,z} = \mathrm{P}(z \mid d)$ where $\mathrm{P}(z \mid d)$ is the probability of the topic $z$ given the document $d$. 

We also represent each query by a probability distribution vector over topics. 
Let $\mathcal{D}_q = \{d_1, d_2, ..., d_n\}$ be the set of top $n$ ranked documents returned for a query $q$ (here, $n=10$). The $z^{th}$ element of the vector embedding for query $q$ is  defined as in \citep{vu2017search}: $\boldsymbol{v}_{q,z} = \sum\nolimits_{i=1}^{n} \lambda_i \mathrm{P}(z \mid d_i)$, where $\lambda_i = \frac{\delta^{{i} - 1}}{\sum_{j=1}^{n}\delta^{{j} - 1}}$ is the exponential decay function of $i$ which is the rank of $d_i$ in $D_q$. And $\delta$ is the decay hyper-parameter ($0 < \delta < 1$). Following \citet{vu2017search}, we use $\delta = 0.8$.
Note that if we learn query and document embeddings during training, the models will overfit to the data and will not work for new queries and documents. 
Thus, after the initialization process, we fix (i.e., not updating) query and document embeddings during training for TransE, ConvKB and CapsE.

In addition, as mentioned by \citet{Bennett2012}, the more recently clicked document expresses more about the user current search interest. Hence, we make use of the user clicked documents in the training set with the temporal weighting scheme proposed by \citet{Vu2015} to initialize user profile embeddings for the three embedding models.

\noindent\textbf{Hyper-parameter tuning:} 
For our CapsE model, we set batch size to 128, and also the number of neurons within the capsule in the second capsule layer to 10 ($d = 10$). The number of iterations in the routing algorithm is set to 1 ($m = 1$).
For the training model, we use the Adam optimizer with the initial learning rate $\in$ $\{5e^{-6},$ $1e^{-5},$ $5e^{-5},$ $1e^{-4},$ $5e^{-4}\}$. We also use ReLU as the activation function $g$. 
We select the number of filters $\mathsf{N} \in \{50, 100, 200, 400, 500\}$.
We run the model up to 200 epochs and perform a grid search to choose optimal hyper-parameters on the validation set.
We monitor the MRR score after each training epoch and obtain the highest MRR score on the validation set when using $\mathsf{N} = 400$ and the initial learning rate at $5e^{-5}$. 

We employ the TransE and ConvKB implementations provided by \citet{NguyenNAACL2016} and \citet{Nguyen2018} and then follow their training protocols to tune hyper-parameters for TransE and ConvKB, respectively.
We also monitor the MRR score after each training epoch and attain the highest MRR score on the validation set when using 
margin = 5, $l_1$-norm and SGD learning rate at $5e^{-3}$ for TransE; and 
$\mathsf{N} = 500$
and the Adam initial learning rate at $5e^{-4}$ for ConvKB. 

\subsection{Main  results}
Table \ref{tab:resultssp} presents the experimental results of the baselines and our model. 
Embedding models TranE, ConvKB and CapsE produce better ranking performances than  traditional learning-to-rank search personalization models CI and SP.
This indicates a prospective strategy of expanding the triple embedding models to improve the ranking quality of the search personalization systems. 
In particular, our MRR and Hits@1 scores are higher than those of TransE (with relative improvements of 14.5\% and 22\% over TransE, respectively). 
Specifically, our CapsE achieves the highest performances in both MRR and Hits@1 (our improvements over all five baselines are statistically significant with $p < 0.05$ using the \textit{paired t-test}).

\begin{table}[!t]
\centering
\begin{tabular}{l|ll}
\hline
\textbf{Method} & \textbf{MRR} & \textbf{H@1}\\
\hline
SE [$\star$]  & 0.559 & 38.5 \\
CI [$\star$]  & 0.597 & 41.6 \\
SP [$\star$]  & 0.631 & 45.2 \\
TransE [$\star$]  & 0.645 & 48.1 \\
\hline
TransE (ours) & 0.669 & 50.9 \\
ConvKB & 0.750$_{+12.1\%}$ & 59.9$_{+17.7\%}$ \\
\hline
Our CapsE & \textbf{0.766}$_{+14.5\%}$ & \textbf{62.1}$_{+22.0\%}$ \\
\hline
\end{tabular}
\caption{Experimental results on the test set. [$\star$] denotes the results reported in  \citep{vu2017search}. Hits@1  (H@1) is reported in \%. In information retrieval, Hits@1 is also referred to as P@1. The subscripts denote the relative improvement over our TransE results.} 
\label{tab:resultssp}
\end{table}

\begin{figure}[!t]
\centering
\includegraphics[width=0.475\textwidth]{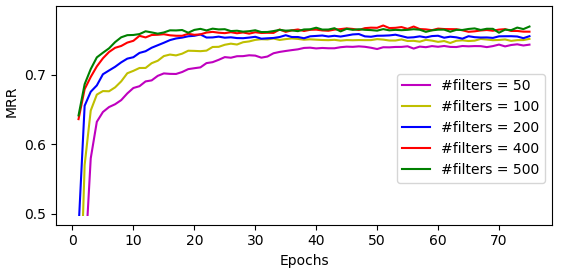}
\captionof{figure}{Learning curves on the validation set with the initial learning rate at $5e^{-5}$.}
\label{fig:learningcurves}
\end{figure}

To illustrate our training progress, we plot performances of CapsE on the validation set over epochs in Figure \ref{fig:learningcurves}. 
We observe that the performance is improved with the increase in the number of filters since capsules can encode more useful properties for a large embedding size.

\section{Related work}

Other transition-based models extend TransE to additionally use projection vectors or matrices to translate embeddings of $s$ and $o$ into the vector space of $r$, such as: TransH \citep{AAAI148531}, TransR \citep{AAAI159571}, TransD \citep{ji-EtAl:2015:ACL-IJCNLP} and STransE \citep{NguyenNAACL2016}. 
Furthermore, DISTMULT \citep{Yang2015} and ComplEx \citep{Trouillon2016} use a tri-linear dot product to compute the score for each triple. 
Moreover, ConvKB \citep{Nguyen2018} applies convolutional neural network, in which 
feature maps are concatenated into a single feature vector which is then computed with a weight vector via a dot product to produce the score for the input triple. ConvKB is the most closely related model to our CapsE. See  an overview of embedding models for KG completion in \citep{Nguyen2017}.  

For search tasks, unlike classical methods, personalized search systems utilize the historical interactions  between the user and the search system, such as submitted queries and clicked documents to tailor returned  results to the need of that user \citep{Teevan2005,Teevan2009}. 
That historical information can be used to build the \emph{user profile}, which is crucial to an effective search personalization system. 
Widely used approaches consist of two separated steps: (1) building the user profile from the interactions between the user and the search system; and then (2) learning a ranking function to \emph{re-rank} the search results using the user profile \citep{Bennett2012,WhiteE2013,HarveyB2013,Vu2015}. 
The general goal is to re-rank the documents returned by the search system in such a way that the more relevant documents are ranked higher.
In this case, apart from the user profile, dozens of other features have been proposed as the input of a learning-to-rank algorithm \citep{Bennett2012,WhiteE2013}. 
Alternatively, \citet{vu2017search} modeled the potential  \textit{user}-oriented relationship between the submitted query and the returned document by applying TransE to reward higher scores for more relevant documents (e.g., clicked documents). They achieved better performances than the standard ranker as well as competitive search personalization baselines \citep{Teevan2011,Bennett2012,Vu2015}.

\section{Conclusion}
\label{sec:conclusion}

We propose CapsE---a novel embedding model using the capsule network to model relationship triples for knowledge graph completion and search personalization.
Experimental results show that our CapsE outperforms other state-of-the-art models on two benchmark datasets WN18RR and FB15k-237 for the knowledge graph completion.
We then show the effectiveness of our CapsE for the search personalization, in which CapsE outperforms the competitive baselines on the dataset SEARCH17 of the web search query logs. 
In addition, our CapsE is capable to effectively model many-to-many relationships. 
Our code is available at: \url{https://github.com/daiquocnguyen/CapsE}.

\section*{Acknowledgement}
This research was partially supported by the ARC Discovery Projects DP150100031 and DP160103934.
The authors thank Yuval Pinter for assisting us in running his code.

\bibliography{references}
\bibliographystyle{acl_natbib}

\end{document}